\def\year{2022}\relax
\documentclass[letterpaper]{article} 
\usepackage{aaai22}  
\usepackage{times}  
\usepackage{helvet}  
\usepackage{courier}  
\usepackage[hyphens]{url}  
\usepackage{graphicx} 
\usepackage{natbib}  
\usepackage{caption} 
\DeclareCaptionStyle{ruled}{labelfont=normalfont,labelsep=colon,strut=off} 
\usepackage{algorithm}
\usepackage{algorithmic}
\usepackage{subfigure}

\usepackage{latexsym}
\usepackage{xcolor}
\usepackage{amsmath}
\usepackage{multirow}
\usepackage{booktabs}
\usepackage{microtype}
\usepackage{stfloats}
\usepackage{amsfonts,amssymb}
\usepackage{bbm}
\usepackage{hyperref}
\usepackage{caption}
\usepackage{color}

\usepackage{newfloat}
\usepackage{listings}
\floatstyle{ruled}
\newfloat{listing}{tb}{lst}{}
\floatname{listing}{Listing}
\title{MvSR-NAT: Multi-view Subset Regularization for Non-Autoregressive \\ Machine Translation}

\author {
    Pan Xie\textsuperscript{\rm 1},
    Zexian Li\textsuperscript{\rm 1},
    Xiaohui Hu\textsuperscript{\rm 2}
}
\affiliations {
    \textsuperscript{\rm 1} Beihang University\\
    \textsuperscript{\rm 2} Chinese Academy of Sciences\\
    panxie@buaa.edu.cn,lizexian0427@gmail.com,hxh@iscas.ac.cn
}

\usepackage{bibentry}

\begin{document}

\maketitle

\begin{abstract}
Conditional masked language models (CMLM) have shown impressive progress in non-autoregressive machine translation (NAT). They learn the conditional translation model by predicting the random masked subset in the target sentence. Based on the CMLM framework, we introduce Multi-view Subset Regularization (MvSR), a novel regularization method to improve the performance of the NAT model. Specifically, MvSR consists of two parts: (1) \textit{shared mask consistency}: we forward the same target with different mask strategies, and encourage the predictions of shared mask positions to be consistent with each other. (2) \textit{model consistency}, we maintain an exponential moving average of the model weights, and enforce the predictions to be consistent between the average model and the online model. Without changing the CMLM-based architecture, our approach achieves remarkable performance on three public benchmarks with 0.36-1.14 BLEU gains over previous NAT models. Moreover, compared with the stronger Transformer baseline, we reduce the gap to 0.01-0.44 BLEU scores on small datasets (WMT16 RO$\leftrightarrow$EN and IWSLT DE$\rightarrow$EN).
\end{abstract}

\section{Introduction}
\label{intro}
\begin{figure}[t]
    \centering
    \subfigure[Vanilla NAT]{
        \includegraphics[width=0.22\textwidth]{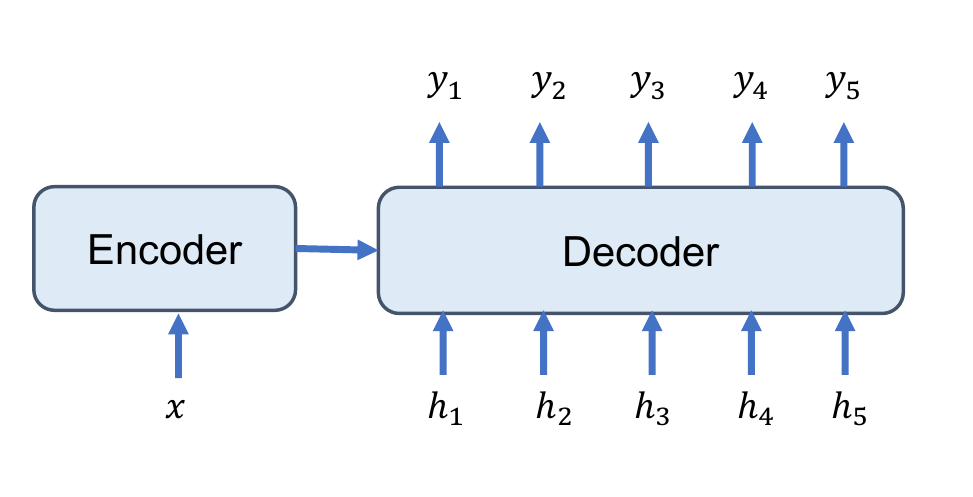}
        \label{vanilla-nat}
    }
    \subfigure[CMLM-NAT]{
	\includegraphics[width=0.22\textwidth]{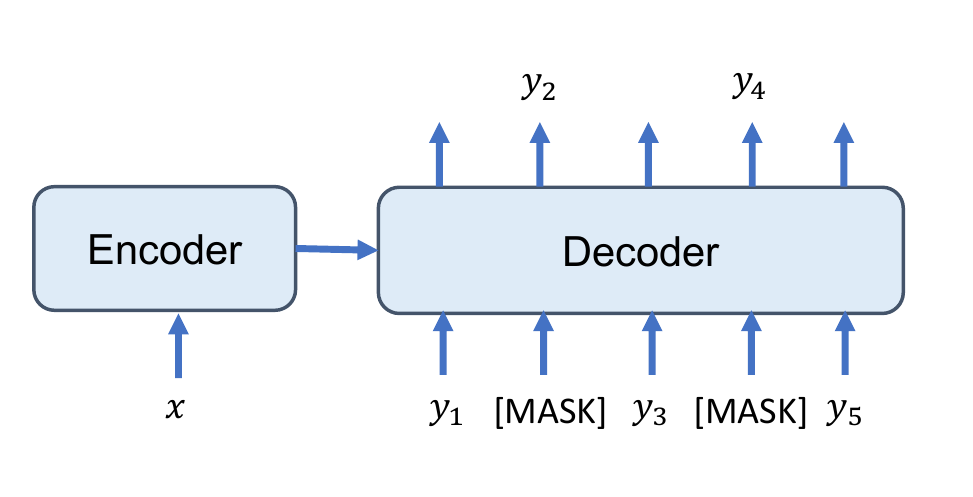}
	\label{cmlm-nat}
    }
    \subfigure[CMLM + \textit{shared mask cons.}]{
	\includegraphics[width=0.22\textwidth]{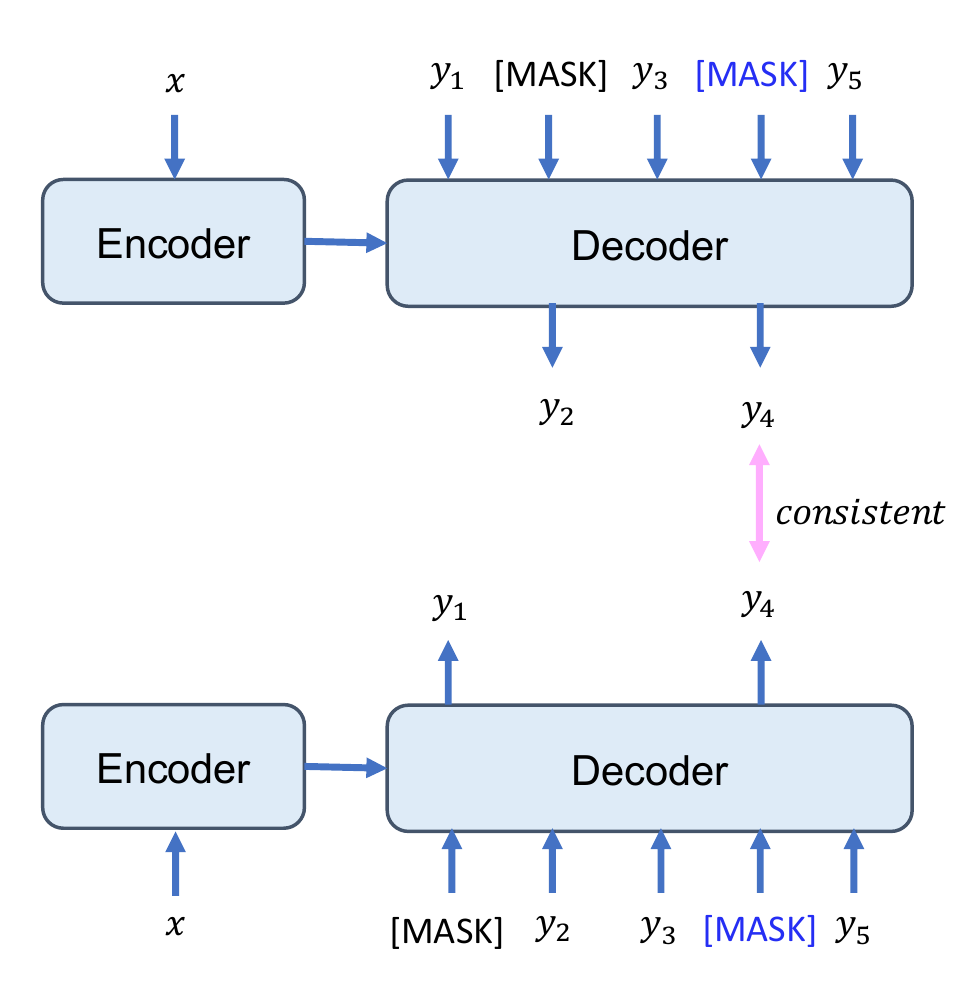}
	\label{cmlm-share-mask}
    }
    \subfigure[CMLM + \textit{model cons.}]{
	\includegraphics[width=0.22\textwidth]{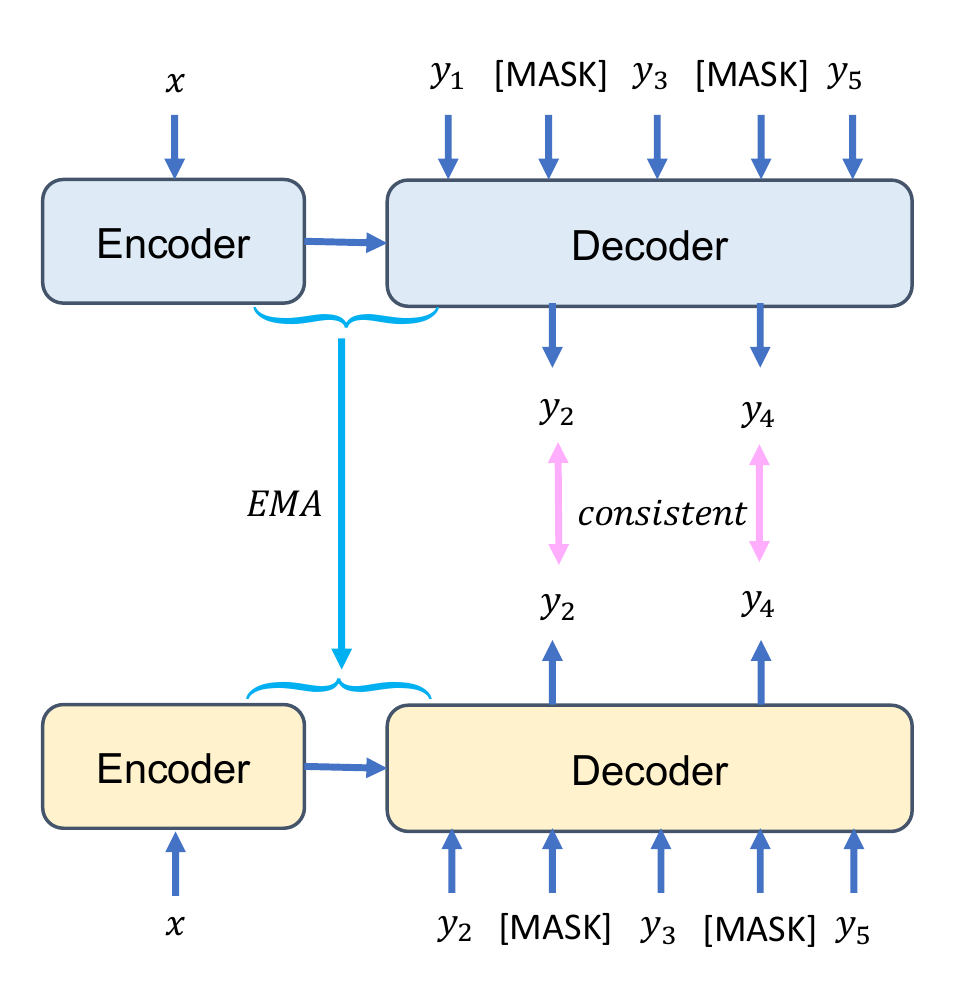}
	\label{cmlm-model-consis}
    }
    \caption{(a) Vanilla NAT model. (b) CMLM-based NAT model. (c) CMLM architecture with shared mask consistency, where the blue \textcolor{blue}{[MASK]} means shared mask position in the two masked target sentences. (d) CMLM architecture with model consistency, where EMA means the  exponential moving average method.}
    \label{arch}
\end{figure}

\noindent Transformer has been the de facto architecture for Neural Machine Translation~\cite{Vaswani2017AttentionIA}. In this framework, the decoder generates words one by one in a left-to-right manner. Despite its strong performance, the autoregressive decoding method causes a large latency in the inference phase~\cite{Gu2018NonAutoregressiveNM}. To break the bottleneck of the inference speed caused by the sequential conditional dependence, several non-autoregressive neural machine translation (NAT) models are proposed to generate all tokens in parallel (Figure~\ref{vanilla-nat})~\cite{Gu2018NonAutoregressiveNM,Kaiser2018FastDI,Li2019HintBasedTF,Ma2019FlowSeqNC}. However, vanilla NAT models suffer from the cost of translation accuracy due to they remove the conditional dependence between target tokens.

To close the gap from autoregressive models, iterative NAT models are proposed to refine the translation results. They bring conditional dependency between target tokens within several iterations~\cite{ghazvininejad2019mask,Ghazvininejad2020SemiAutoregressiveTI,Xie2020InfusingSI,kasai2020non,Guo2020JointlyMS}. Among them, Ghazvininejad~\textit{et al.}~\shortcite{ghazvininejad2019mask} first explore to apply conditional masked language model (CMLM) on NAT model (Figure~\ref{cmlm-nat}). Following this framework, several CMLM-based NAT models are proposed and obtain state-of-the-art performance compared with other NATs~\cite{Xie2020InfusingSI,Guo2020JointlyMS}. 

An open question is whether the potential of the CMLM-based NAT model has been fully exploited, since the masked language model has achieved significant breakthroughs in natural language processing.

To answer the question, we introduce Multi-view Subset Regularization (MvSR), a novel regularization method to improve the performance of the CMLM-based NAT model. Specifically, our approach includes two regularization methods: \textit{shared mask consistency} and \textit{model consistency}. For \textit{shared mask consistency}, as shown in Figure~\ref{cmlm-share-mask}, we randomly mask different subset of the same target sentence twice. Then we encourage the predicted distributions of the shared masked positions to be consistent with each other.  As one example, consider the original sentence and two masked sentences in Table~\ref{mask-reg}. The original token "window" is replaced with \textcolor{blue}{[MASK]} in both two masked sentences. Although the contexts of "window" are different due to the random mask strategies, their semantics and generated distributions are expected to be consistent across these two views. To make a summary, we introduce a new paradigm of 
regularization, different mask strategies for the same target sentence, and the tokens on the shared masked positions are semantic-preserving with different views. This approach is reminiscent of multi-view contrast learning~\cite{Tian2020ContrastiveMC}, our method is not "contrast" but only considers the consistency of "positive pairs".

\begin{table}[t]
\renewcommand\arraystretch{1.2}
\centering
\resizebox{0.48\textwidth}{0.025\textheight}{
\begin{tabular}{c|ccccccccccc}
\hline
original & the & cat & went & through & an & open & window & in & the & house & .\\ 
masked & the & cat & [MASK] & [MASK] & an & open & \textcolor{blue}{[MASK]} & in & the & house & .\\
masked & the & cat & went & through & an & [MASK] & \textcolor{blue}{[MASK]} & in & the & [MASK] & .\\
\hline
\end{tabular}}
\caption{An example target sentence is randomly masked twice. The blue \textcolor{blue}{[MASK]} indicates that the token is masked in both two masked sentences.}
\label{mask-reg}
\end{table}

Regarding \textit{model consistency} (Figure~\ref{cmlm-model-consis}), it is inspired by that checkpoint averaging is an essential method for improving the performance of machine translation~\cite{Vaswani2017AttentionIA}. Similarly, Mean Teacher~\cite{Tarvainen2017MeanTA} shows that using an average model as a teacher improves the results. Correspondingly, we construct an average model by updating the weights with an exponential moving average (EMA) method. Then we penalize the generated distributions that are inconsistent between this average model and the online model. Note that we adopt the bidirectional Kullback-Leibler (KL) divergence instead of mean squared error (MSE) as the consistency cost. This is related to mutual learning~\cite{Zhang2018DeepML} but without extra parameters.

As in prior work, we apply our MvSR-NAT model in several public benchmark datasets. It outperforms previous NAT models and achieves comparable results with autoregressive Transformer. Intuitively, our two proposed regularization methods have two advantages: 1) they can be seen as stabilizers to promote the robustness of the model to randomness; 2) they reduce the discrepancy between the training and inference phase. 

Specifically, the \textit{shared mask consistency} first enhances the robustness of the model to the random mask. Secondly, we adopt the mask-predict decoding method~\cite{ghazvininejad2019mask}, where the predicted target tokens are replaced by [MASK] symbols in the inference process. Especially in the first decoder iteration, all the target tokens are [MASK] symbols. This decoding strategy causes the discrepancy from training for random mask. Therefore, As a result of a more robust model to random mask, our proposed method can reduce the discrepancy between training and inference caused by [MASK] symbols, thus improving the translation quality. 

As for \textit{model consistency}, it first penalizes the model sensitivity to the model weights, thus improving the robustness. Secondly, the average model and the online model have the same architecture but with different dropout units during training. Therefore, this regularization item also makes our model more robust to random dropout. Moreover, the dropout is closed during inference thus causing the discrepancy between training and inference. By reason of more robust to dropout, the proposed \textit{model consistency} method implicitly strengthens the generalization ability of the model and improves the performance with dropout closing during inference. 


Experimental results demonstrate that our model outperforms several state-of-the-art NAT models by over 0.36-1.14 BLEU on WMT14 EN$\leftrightarrow$DE and WMT16 EN$\leftrightarrow$RO datasets. Compared with the strong autoregressive Transformer (AT) baseline, our proposed NAT model achieves competitive performance, while significantly reducing the cost of time during inference.

\section{Background}
\subsection{Non-autoregressive Machine Translation}

Recently, we have witnessed tremendous progress in neural machine translation (NMT)~\cite{Sutskever2014SequenceTS,Bahdanau2015NeuralMT,Vaswani2017AttentionIA}. Given a source sentence $X=\{x_1, x_2,..,x_M\}$, a NMT model is aimed to generate target sentence $Y=\{y_1,y_2,...,y_N\}$. Typically, the probability model of an autoregressive model is defined as $P_{at}(Y|X;\theta)$, where $\theta$ is the parameters of a network. It is formulated as a chain of conditional probabilities:

\begin{equation}
\begin{aligned}
{P_{at}(Y|X;\theta)} = \prod_{t=1}^{N+1}{P(y_t|y_{0:t-1}, X;\theta)}
\end{aligned}
\label{eqn:equation1}
\end{equation}

\noindent where $y_0$ and $y_{N+1}$ are [BOS] and [EOS], representing the beginning and the end of a target sentence, respectively. Note that the autoregressive NMTs adopt teacher forcing~\cite{Vaswani2017AttentionIA} method to capture the sequential conditional dependency between target tokens. And during inference, they generate the target tokens one by one in a left-to-right manner.

As the performance of NMT models have been substantially promoted, non-autoregressive machine translation (NAT) with paralleled decoding becomes a research hotspot~\cite{Gu2018NonAutoregressiveNM}, the architecture is shown in Figure~\ref{vanilla-nat}. We define the probability model of a NAT model as $P_{nat}(Y|X;\theta)$. Mathematically, it is parameterized by a conditional independent factorization:

\begin{equation}
\begin{aligned}
P_{nat}(Y|X)=\prod_{t=1}^{N}P(y_t|X;\theta)
\end{aligned}
\label{eqn:equation2}
\end{equation}

NAT model removes the conditional dependency between the target words, thus generating all target tokens simultaneously. Although the translation speed has been significantly accelerated, the lack of dependence between target words reduces the translation quality. To promote the performance of NAT models, a promising research line is iterative decoding methods~\cite{lee2018deterministic,ghazvininejad2019mask,Ghazvininejad2020SemiAutoregressiveTI,Xie2020InfusingSI,kasai2020non,Guo2020JointlyMS}. Specifically, they explicitly consider the conditional dependency between target tokens within several decoding iterations, thus refining the translation results.

\subsection{Conditional Masked Language Model}
Since the masked language model (MLM) is proposed by BERT~\cite{Devlin2019BERTPO}, it has achieved a significant breakthrough in natural language understanding~\cite{Liu2019RoBERTaAR,Joshi2020SpanBERTIP,Song2019MASSMS,Dong2019UnifiedLM,Sun2021ERNIE3L,Ahmad2021UnifiedPF}. However, due to the bidirectional nature of MLM, it is non-trivial to extend MLM for language generation tasks. Wang~\textit{et al.}~\shortcite{Wang2019BERTHA} start with a sentence of all [MASK] tokens and generate words one by one in arbitrary order (instead of the standard left-to-right chain decomposition), obtaining inadequate generation quality compared with autoregressive counterparts~\cite{Brown2020LanguageMA}. Recently, XLM~\cite{Lample2019CrosslingualLM} leverages sentence-pair translation data for training a conditional masked language model (CMLM), which improves the performance on several downstream tasks, including machine translation. 

Upon previous works, as shown in Figure~\ref{cmlm-nat}, Ghazvininejad~\textit{et al.}~\cite{ghazvininejad2019mask} adopt CMLM to optimize the non-autoregressive NMT model. During training, they predict the masked target tokens $Y_{mask}$ conditional on the source sentence $X$ and the rest of observed words $Y_{obs}$ in the target sentence. Therefore, the training objective of the CMLM-based NAT is presented as:

\begin{equation}
\begin{aligned}
P_{cmlm-nat}(Y_{mask}|X)=\prod_{t=1}^{|Y_{mask}|}P(y_{mask}^{t}|X,Y_{obs};\theta)
\end{aligned}
\label{eqn:equation3}
\end{equation}

\noindent where $|Y_{mask}|$ denotes the number of masked tokens in target sentence. During inference, they propose a Mask-Predict decoding strategy, which iteratively refines the generated translation given the most confident target words predicted from the previous iteration. In this paper, our work is built upon this CMLM-based NAT model and improve its performance with two proposed consistency regularization techniques.

\subsection{Consistency Regularization}
Consistency regularization has merged as a gold-standard technique for semi-supervised learning~\cite{Sajjadi2016RegularizationWS,Laine2017TemporalEF,Zhai2019S4LSS,Oliver2018RealisticEO,Xie2020UnsupervisedDA,Zheng2021ConsistencyRF}. One strand of this idea is to regularize predictions to small perturbations on image data or language. These semantic-preserving augmentations can be image flipping or cropping, or adversarial noise on image data~\cite{8417973,Carmon2019UnlabeledDI}  and natural language example~\cite{zhu2019freelb,Liu2020AdversarialTF}. Another strand of consistency regularization aims at penalizing sensitivity to model parameters~\cite{Tarvainen2017MeanTA,Athiwaratkun2019ThereAM,Liang2021RDropRD}. In our work, we focus on the conditional masked language model setting, leveraging the two strands of consistency regularization.

\begin{figure*}[t]
\centering
\includegraphics[width=0.98\textwidth]{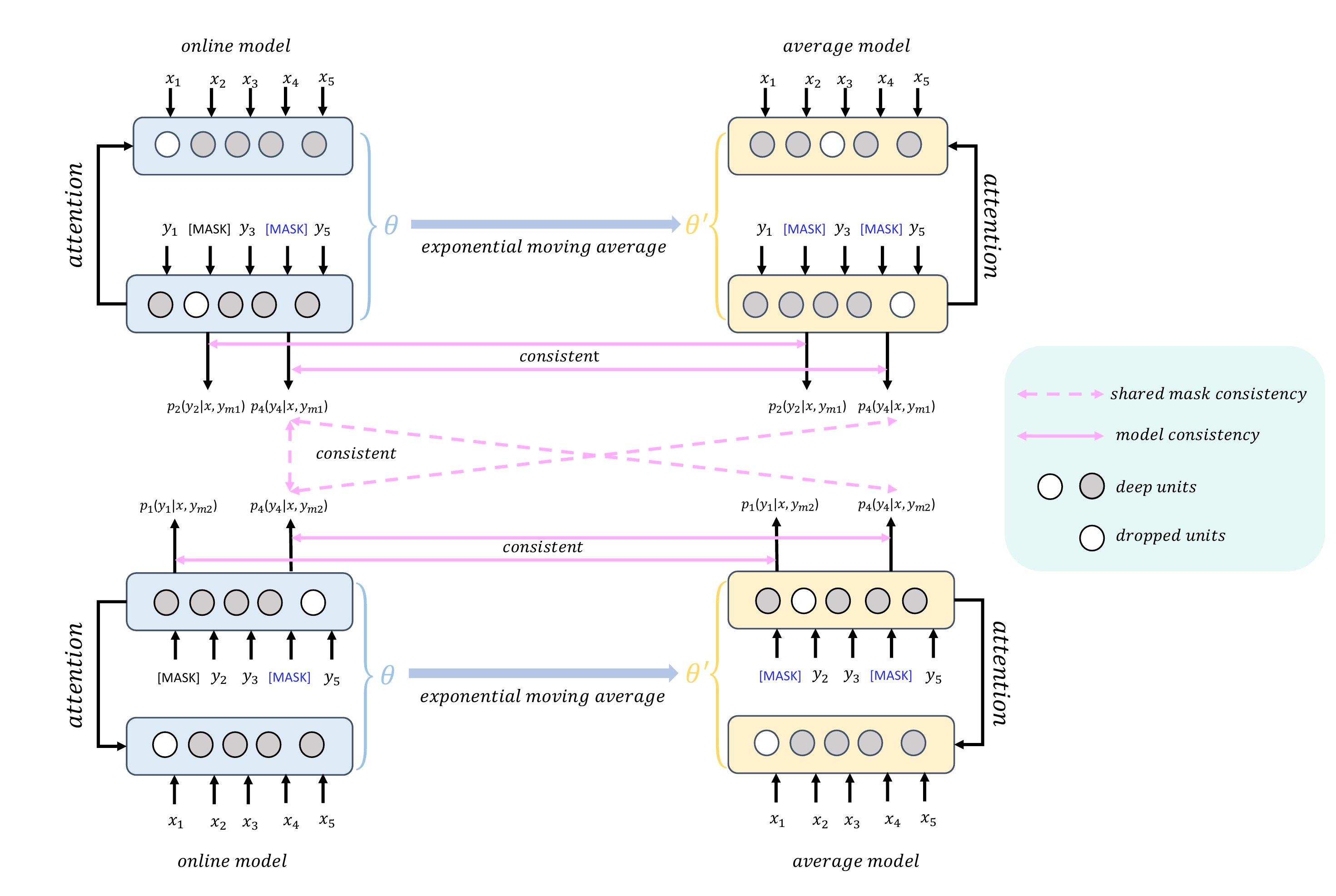}
\caption{The framework of our MvCR-NAT model, consisting of an online model and an average model. This figure depicts a training example with two different mask strategy. Both the two masked sentences are feed to these two models. We update the parameters of the online model with gradient descent.Then we update the average model with an exponential moving average method after every training step.}
\label{mvcr-nat} 
\end{figure*}

\section{Approach}
\subsection{Model Architecture}

Figure~\ref{mvcr-nat} illustrates the overall architecture of our proposed MvCR-NAT model. It is built upon the CMLM framework, comprising of two Transformer-based modules, an Encoder and a Decoder. It is worth noting that the Encoder structure is based on Transformer~\cite{Vaswani2017AttentionIA}, while the Decoder is slightly different. Specifically, the Decoder replaces the left-to-right mask with a bidirectional attention mask, allowing the Decoder to leverage both left and right contexts to predict the target words. To focus on our main contributions, we omit the detailed architecture and refer readers to ~\cite{ghazvininejad2019mask} for reference.


In our work, we focus on the training of our model with two proposed regularization methods. Before elaborating on the details, we first present some notations. Given a source sentence $X=\{x_1, x_2,..,x_M\}$ and a target sentence $Y=\{y_1,y_2,...,y_N\}$, the goal of training is to learn a probability model $P(Y|X;\theta)$. To construct the target input for our approach, we randomly mask the target sentence twice. We define the subset of masked tokens as $Y_{ms1}$ and $Y_{ms2}$, and the observed unmasked tokens as $Y_{obs1}$ and $Y_{obs2}$, respectively. As shown in Figure~\ref{mvcr-nat}, we feed the two masked sentences to the online model (blue part in the figure). Upon the probability model of the CMLM-based NAT presented in Equation~\ref{eqn:equation3}, our learning objective is learning to minimize the negative log-likelihood (NLL) loss of the masked tokens, which is parameterized as:

\begin{equation}
\begin{aligned}
\mathcal{L}_{nll}^{1} &=-\sum_{t=1}^{|Y_{ms1}|}log P_o(y_{ms1}^{t}|X,Y_{obs1};\theta) \\
\mathcal{L}_{nll}^{2} &=-\sum_{t=1}^{|Y_{ms2}|}log P_o(y_{ms2}^{t}|X,Y_{obs2};\theta)
\end{aligned}
\label{eqn:equation4}
\end{equation}

\noindent where $\theta$ represents the parameters of the online model. $|Y_{ms1}|$ and $|Y_{ms2}|$ represent the number of masked words, respectively. $P_o(y_{ms1}^{t}|X,Y_{obs1};\theta)$ and $P_o(y_{ms2}^{t}|X,Y_{obs2};\theta)$ represent the generated distributions from the online model. And we obtain the two nll losses $\mathcal{L}_{nll}^{1}$ and $\mathcal{L}_{nll}^{2}$, respectively.

Note that the masked words in $Y_{ms1}$ and $Y_{ms2}$ are randomly selected and replaced by the [MASK] symbols. As shown in Table~\ref{mask-reg}, the shared masked tokens in both $Y_{ms1}$ and $Y_{ms2}$ are marked as \textcolor{blue}{[MASK]} symbol. And we set the collection of the shared masked words as $Y_{s-ms} = Y_{ms1} \cap Y_{ms2}$.

\subsection{Consistency Regularization}
In this Section, we propose to improve the CMLM-based NAT with consistency regularization method. Specifically, we focus on regularizing the generated predictions to be invariant to model parameters and semantic-preserving data perturbations. 

\subsubsection{Model Consistency}\label{model_consis}
We introduce the \textit{model consistency} regularization method to encourage consistent predictions from an online model and an average model. The average model weights are maintained by an exponential moving average (EMA) method (yellow part in Figure~\ref{mvcr-nat}). Previous works have demonstrated that averaging model weights tend to achieve better performance than using the final model weight~\cite{polyak1992acceleration,Vaswani2017AttentionIA,Tarvainen2017MeanTA}. To take this advantage of the average model, we adopt the bidirectional KL divergence to encourage the prediction consistency between these two models. Therefore we can increase the robustness of our model to model weights and learn better representations. Furthermore, similar to the recent ARXIV paper~\cite{Liang2021RDropRD}, the dropout units between the online model and the average model is different due to the randomness. Thus the prediction consistency penalizes the sensitivity to random dropout. Totally, the proposed method brings two practical advantages: first, we strengthen the robustness of our model to stochastic model weights. Second, this method robustly improves the model generalization and reduces the discrepancy between training and inference caused by dropout.

Formally, we define the model consistency loss $\mathcal{L}_{m-kl}$ as the distance between the token-level predictions produced by the online model and the average model using the bidirectional KL divergence:

\begin{equation}
\begin{aligned}
\mathcal{L}_{m-kl}^{1} = &\dfrac{1}{|Y_{ms1}|}\sum_{t=1}^{|Y_{ms1}|} [\dfrac{1}{2}(D_{kl}(P_o(y_{ms1}^t;\theta)||P_a(y_{ms1}^t;\theta^{'})) \\ & + D_{kl}(P_a(y_{ms1}^t;\theta^{'})||P_o(y_{ms1}^t;\theta)))] \\
\mathcal{L}_{m-kl}^{2} = &\dfrac{1}{|Y_{ms2}|}\sum_{t=1}^{|Y_{ms2}|} [\dfrac{1}{2}(D_{kl}(P_o(y_{ms2}^t;\theta)||P_a(y_{ms2}^t;\theta^{'})) \\ & + D_{kl}(P_a(y_{ms2}^t;\theta^{'})||P_o(y_{ms2}^t;\theta)))]
\end{aligned}
\label{eqn:equation6}
\end{equation}

\noindent where $\theta^{'}$ represents the parameters of the average model. In order to ensure readability, $P_o(y_{ms1}^t;\theta)$ and $P_a(y_{ms1}^t;\theta^{'})$ are the abbreviation of $P_o(y_{ms1}^{t}|X,Y_{obs1};\theta)$ and $P_a(y_{ms1}^{t}|X,Y_{obs1};\theta^{'})$. They represent the predictions from the online model and the average model with the first masked sentence, respectively. Similar for $P_o(y_{ms2}^t;\theta)$ and $P_a(y_{ms2}^t;\theta^{'})$, but with the second masked sentence.

Moreover, the average model parameters $\theta^{'}$ is obtained by EMA method. At training step $t$, the updated $\theta^{'}_{t}$ is computed as the EMA of successive $\theta$ weights:

\begin{equation}
\begin{aligned}
\theta^{'}_{t} = \alpha \theta^{'}_{t-1} + (1-\alpha)\theta_t
\end{aligned}
\label{eqn:equation5}
\end{equation}

\subsubsection{Shared Mask Consistency}

We propose our shared mask consistency regularization method in this part. As examples shown in Table~\ref{mask-reg} and Figure~\ref{mvcr-nat}, we randomly mask the same target sentence twice, and forward them to the online model and the average model. Considering a simple example, there is a sentence pairs "\textit{the [MASK] is \textcolor{blue}{[MASK]} . $|$ Diese Katze ist lustig .}" and "\textit{[MASK] cat is \textcolor{blue}{[MASK]} . $|$ Diese Katze ist lustig .}". The shared \textcolor{blue}{[MASK]} to predict can be thought of a token-level "positive pair" with semantic-preserving. We hypothesis that the representation is view-agnostic, and the semantic is shared between different views caused by randomly mask. Therefore, the predicted distributions of shared masked tokens are expected to be consistent. Note that we do not consider the distribution consistency of the other positions. To illustrate this reason, let’s take an example, in the second position, "cat" is observed in the second target sentence, but it is masked in the first sentence. If we force the distributions of the second position to be consistent, the model will be confused with "[MASK]" and "cat", thus leading inferior performance.

Mathematically, we define the shared mask consistency cost $\mathcal{L}_{s-kl}$ to measure the distance of prediction distributions in shared mask position. similar to model consistency, we adopt the bidirectional KL divergence:

\begin{equation}
\begin{aligned}
\mathcal{L}_{s-kl}^{1} = &\dfrac{1}{|Y_{s-ms}|}\sum_{t=1}^{|Y_{s-ms}|} [\dfrac{1}{2}(D_{kl}(P_o(y_{ms1}^t;\theta)||P_o(y_{ms2}^t;\theta)) \\ & + D_{kl}(P_o(y_{ms2}^t;\theta)||P_o(y_{ms1}^t;\theta)))] \\
\mathcal{L}_{s-kl}^{2} = &\dfrac{1}{|Y_{s-ms}|}\sum_{t=1}^{|Y_{s-ms}|} [\dfrac{1}{2}(D_{kl}(P_o(y_{ms1}^t;\theta)||P_a(y_{ms2}^t;\theta^{'})) \\ & + D_{kl}(P_a(y_{ms2}^t;\theta^{'})||P_o(y_{ms1}^t;\theta)))] \\
\mathcal{L}_{s-kl}^{3} = &\dfrac{1}{|Y_{s-ms}|}\sum_{t=1}^{|Y_{s-ms}|} [\dfrac{1}{2}(D_{kl}(P_a(y_{ms1}^t;\theta^{'})||P_o(y_{ms2}^t;\theta) \\ & + D_{kl}(P_o(y_{ms2}^t;\theta)||P_a(y_{ms1}^t;\theta^{'}))))] \\
\end{aligned}
\label{eqn:equation7}
\end{equation}

\noindent where $|Y_{s-ms}|$ represents the number of shared masked tokens between $Y_{ms1}$ and $Y_{ms2}$. $\mathcal{L}_{s-kl}^{1}$, $\mathcal{L}_{s-kl}^{2}$, $\mathcal{L}_{s-kl}^{3}$ represent the share mask consistency between the two masked sentence when they are feed into online-online, online-average, and average-online models, respectively.

\subsection{Length Prediction}
Autoregressive NMT models generate words one-by-one, and the length of the target sentence is decided by encountering a special token [EOS]. However, non-autoregressve NATs generate target sentence in a parallel way, thus requiring the predicted length before decoding. Following~\cite{Gu2018NonAutoregressiveNM,ghazvininejad2019mask}, we add an additional special token [LEN] to the begining of the source input. Then we predict the length $L$ by the source sentence X. Mathematically, we define the loss function of this classfication task as:

\begin{equation}
\begin{aligned}
\mathcal{L}_{len} = \sum_{i}^{N_{max}}-(L=i)logP(L|X)
\end{aligned}
\label{eqn:equation8}
\end{equation}

\noindent where $N_{max}$ represents the max length of the target sentence in our corpus.

\begin{algorithm}[tb]
\caption{Training Algorithm.}
\label{alg:algorithm1}
\textbf{Input}: Training data $\mathcal{D}=\{X, Y\}^{n}$, $X=\{x_1,x_2,...,x_M\}$, $Y=\{y_1,y_2,...,y_N\}$\\
\textbf{Output}: online model parameters $\theta$ and average model parameters $\theta^{'}$
\begin{algorithmic}[1] 
\STATE Initialize $\theta$ and copy $\theta$ to $\theta^{'}$;\label{code:fram:init}
\WHILE{not converged do}\label{code:fram:startwhile}
\STATE randomly sample data $(X,Y) \sim \mathcal{D}$; \label{code:fram:sampledata}
\STATE randomly mask $Y$ twice, and obtain two masked sentences, where the masked subset are $Y_{ms1}$ and $Y_{ms2}$, the observed subset are $Y_{obs1}$ and $Y_{obs2}$; \label{code:fram:masktarget}
\STATE feed the two sentence pairs to the online model, obtain the distribution $P_o(Y_{ms1}|X, Y_{obs1};\theta)$ and $P_o(Y_{ms2}|X, Y_{obs2};\theta)$;\label{code:fram:online}
\STATE feed the same examples to the average model, obtain the distribution $P_a(Y_{ms1}|X, Y_{obs1};\theta^{'})$ and $P_a(Y_{ms2}|X, Y_{obs2};\theta^{'})$;\label{code:fram:average}
\STATE calculate the log-likelihood loss $\mathcal{L}_{nll}^{1}$ and $\mathcal{L}_{nll}^{2}$ by Equation~\ref{eqn:equation4};\label{code:fram:nll_loss} 
\STATE calculate the model consistency losses $\mathcal{L}_{m-kl}^{1}$ and $\mathcal{L}_{m-kl}^{2}$ by Equation~\ref{eqn:equation6};\label{code:fram:model_loss}
\STATE calculate the shared mask consistency losses $\mathcal{L}_{s-kl}^{1}$, $\mathcal{L}_{s-kl}^{2}$ and $\mathcal{L}_{s-kl}^{3}$ by Equation~\ref{eqn:equation7}; \label{code:fram:mask_loss}
\STATE calculate the length prediction loss $\mathcal{L}_{len}$ by Equation~\ref{eqn:equation8};\label{code:fram:len_loss}
\STATE update the online model parameters $\theta$ by minimizing the total loss $\mathcal{L}$ of Equation~\ref{eqn:equation9};\label{code:fram:optimizer}
\STATE update the average model parameters $\theta^{'}$ using EMA method.
\ENDWHILE
\end{algorithmic}
\end{algorithm}

\subsection{Training Algorithm}


The final training objective for MvSR-NAT is the sum of all aforementioned loss functions:

\begin{equation}
\begin{aligned}
\mathcal{L} = & \dfrac{1}{2}(\mathcal{L}_{nll}^1 + \mathcal{L}_{nll}^2) + \dfrac{\lambda}{5}(\mathcal{L}_{m-kl}^{1} + \mathcal{L}_{m-kl}^{2} + \mathcal{L}_{s-kl}^{1} \\
& + \mathcal{L}_{s-kl}^{2} + \mathcal{L}_{s-kl}^{3}) +\mathcal{L}_{len}
\end{aligned}
\label{eqn:equation9}
\end{equation}

\noindent where $\lambda$ is a hyperparameter to control KL losses. Jointly training with our proposed two consistency regularization losses, we improve the robustness and generalization ability of our MvCR-NAT model to the randomness (e.g., model weights and random mask in target sentence). 

The overall training algorithm of our model is presented in Algorithm~\ref{alg:algorithm1}.

\subsection{Inference}
During inference, we feed a sentence with all [MASK] as target input for the first iteration, where its length is determined by the length prediction. Then we refine the translation result by masking-out and re-predicting a subset of words whose probabilities are under a threshold within several iterations. For more details, please refer to~\cite{ghazvininejad2019mask}.

\section{Experiments}
\subsection{Experimental Setup}
\subsubsection{Datasets}
We conduct our experiments on five public benchmarks: WMT14 EN$\leftrightarrow$DE (4.5M translation pairs), WMT16 EN$\leftrightarrow$RO (610K translation pairs), and IWSLT DE$\rightarrow$EN (150K translation pairs). We strictly follow the dataset configurations of previous works. Specifically, we preprocess the dataset following~\cite{lee2018deterministic}. Then we tokenize the tokens into subword units using BPE method~\cite{sennrich2016neural}. For WMT14 EN$\leftrightarrow$DE, we use newstest-2013 and newstest-2014 as our development and test datasets, respectively. For WMT16 En$\leftrightarrow$Ro, we use newsdev-2016 and newstest-2016 as our development and test datasets, respectively. 

\subsubsection{Evaluation Metrics}
We adopt the widely used BLEU~\cite{Papineni2001BleuAM} to evaluate the translation quality.

\begin{table*}[t]
\renewcommand\arraystretch{1.0}
\centering
\smallskip
\resizebox{1.0\textwidth}{!}{
\begin{tabular}{l|c|ccccc|c}
\toprule[1pt]
\multirow{2}{*}{\textbf{Models}} &\multirow{2}{*}{\textbf{\textbf{$I_{dec}$}}} & \multicolumn{2}{c}{\textbf {WMT'14}} & \multicolumn{2}{c}{\textbf {WMT'16}} & \textbf{IWSLT14}& \multirow{2}{*}{\textbf{Speedup}}  \\
& & \textbf {EN\(\rightarrow\)DE} & \textbf {DE\(\rightarrow\)EN} & \textbf {EN\(\rightarrow\)RO} & \textbf {RO\(\rightarrow\)EN} & \textbf{DE\(\rightarrow\)EN}\\
\midrule
\multicolumn{4}{l}{\textit{\textbf{Autoregressive Models}}} \\
\midrule
\small{LSTMS2S\cite{Bahdanau2015NeuralMT}} & \small{T} &  \small{24.60} & \small{-} & \small{-} & \small{-} & \small{-} & \small{-} \\
\small{ConvS2S\cite{Gehring2017ConvolutionalST}} & \small{T} & \small{26.42} & \small{-} & \small{-} & \small{-} & \small{-} & \small{-} \\
\small{Transformer~\cite{Vaswani2017AttentionIA}} & \small{T} & \small{\underline{28.04}} & \small{\underline{32.69}} & \small{\underline{34.13}} & \small{ \underline{34.46}} &\small{\underline{32.99}} & \small{1.00x} \\
\midrule
\multicolumn{4}{l}{\textit{\textbf{Non-Autoregressive Models}}}  \\
\midrule
\small{NAT-FT~\cite{Gu2018NonAutoregressiveNM}} & \small{1} & \small{19.17} & \small{23.20} & \small{29.79} & \small{31.44} & \small{24.21} & \small{2.36x}\\
\small{Imit-NAT~\cite{wei2019imitation}} & \small{1} & \small{24.15} & \small{27.28} & \small{31.45} & \small{31.81} & \small{-} & \small{-} \\
\small{NAT-Hint~\cite{Li2019HintBasedTF}} & \small{1} & \small{21.11} & \small{25.24} & - & - & - & - \\
\small{Flowseq~\cite{Ma2019FlowSeqNC}} & \small{1} & \small{23.72} & \small{28.39} & \small{29.73} & \small{30.72} & - & \small{1.1x}\\
\small{NAT-DCRF~\cite{sun2019fast}} & \small{1} & \small{26.07} & \small{29.68} & - & - & \small{29.99} & \small{9.63x} \\
\small{GLAT-NAT~\cite{Qian2020GlancingTF}} & \small{1} & \small{26.55} & \small{31.02} & \small{32.87} & \small{33.51} & - & \small{7.9x} \\
\small{NAT-IR~\cite{lee2018deterministic}} & \small{5} & \small{20.26} & \small{23.86} & \small{28.86} & \small{29.72} & - & - \\ 
 & \small{10} & \small{21.61} & \small{25.48} & \small{29.32} & \small{30.19} & \small{23.94} & \small{1.50x} \\
\small{CMLM-NAT ~\cite{ghazvininejad2019mask}} & \small{1} &  \small{18.05} & \small{21.83} & \small{27.32} & \small{28.20} & \small{-} & \small{27.51x} \\
& \small{4} &  \small{25.94} & \small{29.90} & \small{32.53} & \small{33.23} & \small{30.42} & \small{9.79x} \\
& \small{10} & \small{27.03} & \small{30.53} & \small{33.08} &  \small{33.31} & \small{31.71} & \small{3.77x} \\
\midrule
\small{\textbf{MvCR-NAT (w/ kd)}} & \small{4} & \small{26.25} & \small{30.27} & \small{32.76} & \small{32.96} & - & \small{9.79x} \\
& \small{10} & \textbf{\small{27.39}} & \textbf{\small{31.18}} & \small{33.38} &  \small{33.56}  & - & \small{3.77x} \\
\midrule
\small{\textbf{MvCR-NAT (w/o kd)}} & \small{4} & \small{22.89} & \small{26.89} & \small{32.34} & \small{33.60} & \small{30.58} & \small{9.79x} \\
& \small{10} & \small{24.37} & \small{28.90} & \textbf{\small{33.76}} & \textbf{\small{34.45}} & \textbf{\small{32.55}} & \small{3.77x} \\
\bottomrule[1pt]
\end{tabular}
}
\caption{The BLEU scores of our proposed MvCR-NAT model and the baseline models on the WMT14 En-De/De-En, WMT16 En-Ro/Ro-En and IWSLT14 De-En tasks. $I_{dec}$ represents the number of iterations while inference. "kd" represents the sequential level knowledge distillation. In the column of speedup, we adopt seconds/sentence to measure the decoding speed, where Transformer is set as the baseline (beam size = 5).}
\label{main results}
\end{table*}

\subsubsection{Model Details}
The online model setting is strictly following previous works. For the WMT14 EN$\leftrightarrow$DE and WMT16 EN$\leftrightarrow$RO datasets, the model setting is based on the base Transformer~\cite{Vaswani2017AttentionIA}. Specifically, we set the model dimension as 512 and the inner dimension as 2048. The Encoder and the Decoder consist of a stack of 6 Transformer layers. For the smaller IWSLT16 DE$\rightarrow$EN dataset, we follow the configuration of small Transformer that the model dimension and inner dimension are 256 and 1024, respectively. The Encoder and the Decoder consist of a stack of 5 Transformer layers. Besides, we set the max target sentence $N_{max}$ as 1000. 

The average model is built upon the online model with the same architecture. Besides, its model weights are maintained by an exponential moving average method with the moving average decay $\alpha$ set as 0.996. 

During training, we train the model with 2048 tokens per batch on eight GTX 2080Ti GPUs. We use Adam optimizer~\cite{Kingma2015AdamAM} and warmup learning rate schedule. During inference, we set the number of the length candidate as 5 for our NAT model. For a fair comparison, we set the beam size as 5 for the baseline AT model. Moreover, we evaluate the final translation accuracy by averaging 10 checkpoints.

\subsection{Sequential-Level Knowledge Distillation}
Previous works have demonstrated that the effectiveness of sequential-level knowledge distillation on NAT models~\cite{Gu2018NonAutoregressiveNM,lee2018deterministic,Gu2019LevenshteinT,ghazvininejad2019mask,Zhou2020UnderstandingKD}. Following their works, we train our CMCR-NAT model on the distilled corpora, which are produced by a standard left-to-right Transformer model. While previous AT transformers have different performances, we adopt the one used in CMLM-NAT~\cite{ghazvininejad2019mask} which is our primary baseline. In Section~\nameref{kd}, we will identify the effect of knowledge distillation on our model.

\begin{table*}[t]
\renewcommand\arraystretch{1.1}
\centering
\smallskip
\resizebox{0.82\textwidth}{!}{
\begin{tabular}{l|cc|cc}
\hline
\multirow{2}{*}{\textbf{\small{Model Variants}}} & \multicolumn{2}{c|}{\textbf{\small{WMT'16 EN$\rightarrow$RO}}} & \multicolumn{2}{c}{\textbf{\small{WMT'16 RO$\rightarrow$EN}}} \\
& \small{$I_{dec}$=4} & \small{$I_{dec}$=10}& \small{$I_{dec}$=4} & \small{$I_{dec}$=10} \\
\hline
\small{CMLM-NAT} & \small{31.40}  & \small{32.86} & \small{32.87} & \small{33.15} \\
\hline
\small{$\quad$+ model consistency} & \small{31.89 (+0.49)}& \small{33.10(+0.24)} & \small{33.15 (+0.28)} & \small{33.52 (+0.37)} \\
\small{$\quad$+ shared mask concsistency} & \small{32.16 (+0.76)}& \small{33.49(+0.63)} &
\small{33.79 (+0.92)}& \small{34.08 (+0.93)} \\
\hline
\small{MvCR-NAT} & \small{32.34 (+0.94)} & \small{33.76(+0.90)} & \small{33.60(+0.73)} & \small{34.45 (+1.30)} \\
\hline
\end{tabular}}
\caption{Evaluation of model consistency and shared mask consistency on WMT16 EN$\leftrightarrow$RO without knowledge distillation. $I_{dec}$ means decoding iterations.}
\label{tabel_modelvsmask}
\end{table*}

\subsection{Results}
Table~\ref{main results} shows our experimental results on three public datasets. As we move our eyes to the first part in this Table, our model achieves comparable performance with the Transformer model. Notably, on the small dataset WMT16 EN$\leftrightarrow$RO and IWSLT14 DE$\rightarrow$EN, the translation results are only 0.01-0.44 BLEU score behind. 

In the second part of Table~\ref{main results}, compared with pure NAT models with one-shot decoding, the multiple iterative decoding methods achieve noticeable improvements. The same thing happens to the CMLM-based NAT models. This phenomenon is mainly due to the problem of multimodality~\cite{Gu2018NonAutoregressiveNM} that the one-shot decoding hardly considers the left-to-right dependency. While the iterative methods explicitly model the conditional dependency between target tokens within several iterations, thus obtaining better performance.

In contrast with our primary baseline CMLM-NAT model, our model is additionally optimized with two regularization methods without changing the CMLM architecture. Our model outperforms CMLM-NAT with margins from 0.36-1.14 BLEU scores, illustrating the effectiveness of our methods.

\subsubsection{Effect of Sequential-level Knowledge Distillation}\label{kd}

The comparison results for knowledge distillation are shown in Table~\ref{main results}. In terms of the large dataset, i.e., WMT14 EN$\leftrightarrow$DE, our model gains improvements with the sequential-level knowledge distillation. However, the improvements from knowledge distillation are not concurrent on the small dataset, i.e., WMT16 EN$\leftrightarrow$RO. We attribute this phenomenon to the complexity of the data sets~\cite{Zhou2020UnderstandingKD}. The knowledge distillation is able to reduce "modes" (alternative translations for an input) in the training data, thus benefiting the NAT models. We conjecture that a small dataset is likely to contain fewer redundant “modes” than a large-scale dataset. As a result, distillation knowledge is helpful and more efficient on a large dataset than on a small dataset.

\begin{figure}[t]
\centering
\includegraphics[width=0.48\textwidth]{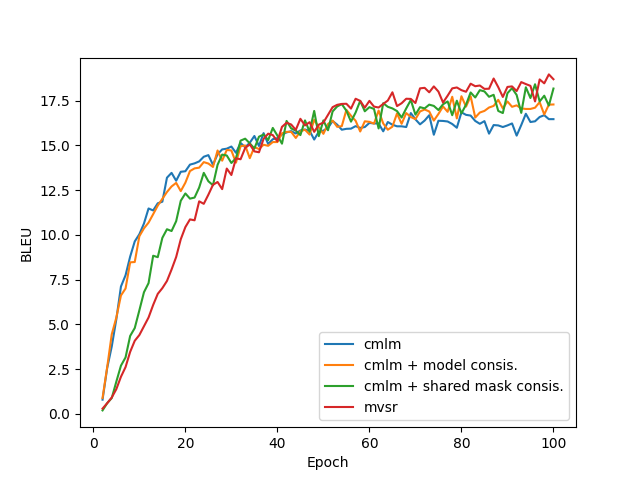}
\caption{The BLEU scores with training epochs on IWSLT14 DE-EN task.}
\label{bleu_epoch} 
\end{figure}

\subsection{Ablation Study}

\subsubsection{Model Consistency vs. Shared Mask Consistency}\label{modelvsmask}

As shown in Table~\ref{tabel_modelvsmask}, we conduct comparative experiments on the validation set of the WMT16 EN$\leftrightarrow$RO task to illustrate the contribution of our proposed two regularization methods. Note that the results are computed without knowledge distillation. Compared with the CMLM-NAT baseline model, our proposed model consistency and shared mask consistency regularization methods progressively improve the performance, and the shared mask consistency provides more performance promotion.

Furthermore, to step further understand the two proposed regularization methods, in Figure~\ref{bleu_epoch}, we show the BLEU score with training epochs on IWSLT14 DE$\rightarrow$EN task with single decoding iteration. To make a fair comparison, in every training forward pass, we feed two source-target sentence pairs to these compared models. The training curves help us understand the effect of the two proposed regularization methods. We can see that a) the model consistency improves the performance without changing the convergence trend; b) the shared mask consistency method suppresses the convergence speed of the model in the early training period, but obtains better performance in the final training epochs. It indicates that the shared mask consistency method can avoid premature fitting and improve the robustness and the generalization ability of our model.

\subsubsection{Effect of Weight $\lambda$}\label{lambda}

\begin{table}[t]
\renewcommand\arraystretch{1.3}
\centering
\smallskip
\resizebox{0.47\textwidth}{!}{
\begin{tabular}{c|c|c}
\hline
\textbf{KL Loss Weight} & \textbf{\small{WMT'16 EN$\rightarrow$RO}} & \textbf{\small{WMT'16 RO$\rightarrow$EN}} \\
\hline
$0.1$ & \small{33.68} & \small{34.40} \\
$0.3$ & \textbf{\small{33.76}} & \textbf{\small{34.45}} \\
$0.5$ & \small{33.42} & \small{34.02} \\
$1$ & \small{33.10}& \small{33.43} \\
$3$ & \small{32.72}& \small{32.68} \\
\hline
\end{tabular}}
\caption{Evaluation of kl loss weight $\lambda$.}
\label{tabel_lambda}
\end{table}

We investigate the effect of the loss weight $\lambda$, which is utilized for controlling the KL-divergence loss. We conduct ablation experiments on WMT16 EN$\rightarrow$RO with different values in $\{0.1, 0.3, 0.5, 1.0, 3.0\}$. The results are shown in Table~\ref{tabel_lambda}. We can see that the small kl loss weight performs better than the larger ones. In our setting, the best choice of the kl loss weight is $\lambda=0.3$. Too much regularization (e.g. 3) even decreases the model performance. 

\subsubsection{$m$-time Shared Mask Consistency}

\begin{table}[t]
\renewcommand\arraystretch{1.2}
\centering
\smallskip
\resizebox{0.47\textwidth}{!}{
\begin{tabular}{c|c|c}
\hline
\textbf{KL Loss Weight} & \textbf{\small{WMT'16 EN$\rightarrow$RO}} & \textbf{\small{WMT'16 RO$\rightarrow$EN}} \\
\hline
$m=2$ & \textbf{\small{33.76}} & \small{34.45} \\
$m=3$ & \small{33.53} & \textbf{\small{34.50}} \\

\hline
\end{tabular}}
\caption{Evaluation of $m$-time Shared Mask Consistency.}
\label{table_msmc}
\end{table}

As an example shown in Table~\ref{mask-reg}, we forward two masked target sentences to the model and encourage their masked subset predictions to be consistent. An interesting concern is whether more improvements can be achieved if we forward three or more masked targets with different mask strategies. In this study, we define the number of masked targets as $m$. We conduct comparative experiments about the $m$ on the WMT16 EN$\leftrightarrow$RO dataset. The results in Table~\ref{table_msmc} show that $m=2$ is good enough for the tasks. This indicates that our proposed two consistency regularization methods have strong regularization effect between two distributions, without the necessity of more distributions regularization.

\subsubsection{Dropout Probability in Average Model}

\begin{table}[t]
\renewcommand\arraystretch{1.2}
\centering
\smallskip
\resizebox{0.48\textwidth}{!}{
\begin{tabular}{c|c|c}
\hline
\textbf{\small{Dropout Prob.}} & \textbf{\small{WMT'16 EN$\rightarrow$RO}} & \textbf{\small{WMT'16 RO$\rightarrow$EN}} \\
\hline
0.1 & \small{33.17} & \small{33.87} \\
0.2 & \small{33.36}& \small{34.19}\\
0.3 & \textbf{\small{33.76}} & \textbf{\small{34.45}} \\
0.4 & \small{33.39} & \small{34.40} \\
0.5 & \small{33.10} & \small{34.25} \\
\hline
\end{tabular}}
\caption{Evaluation of dropout probability in average model.}
\label{tabel_drop}
\end{table}

As mentioned in subsection~\nameref{model_consis}, we indicate that the model consistency method strengthens the robustness of our model to model weights and randomly dropout. Here, we investigate the different dropout rates in the average model. In this study, we apply different dropout values for the average models during training. As shown in Table~\ref{tabel_drop}, we test the dropout values from $\{0.1, 0.2, 0.3, 0.4, 0.5\}$ on WMT16 EN$\leftrightarrow$RO dataset. We can see that the best choice of dropout value for the average model is 0.3.

\section{Conclusion}

In this paper, upon CMLM-based architecture, we introduce the Multi-view Subset Regularization method to improve the CMLM-based NAT performance. We first propose the shared mask consistency method to force the masked subset predictions to be consistent for randomly mask strategies. Second, we propose model consistency to encourage the online model to generate consistent distributions with the average model whose weights are maintained with an EMA method. On several benchmark datasets, we demonstrate that our approach achieves considerable improvements against previous non-autoregressive models and comparable results to the autoregressive Transformer model. This work introduces a new paradigm for regularization method, the multi-view subset regularization. We hope this paradigm can be helpful in recent hot contrast learning models.

\bibliography{aaai22}

\end{document}